\pdfoutput=1
\documentclass[11pt]{article}

\usepackage[]{ACL2023}

\usepackage{times}
\usepackage{latexsym}

\usepackage[T1]{fontenc}
\usepackage[utf8]{inputenc}

\usepackage{microtype}

\usepackage{url}
\usepackage{multirow}
\usepackage{booktabs}
\usepackage{tabularx}
\usepackage{graphics}
\usepackage{graphicx}
\usepackage{colortbl}
\usepackage{xcolor,soul}
\usepackage{makecell}
\usepackage{amsmath}
\DeclareMathOperator*{\argmax}{argmax}

\definecolor{grey}{HTML}{D3D3D3}
\definecolor{berry}{HTML}{E3A3B4}
\definecolor{sage}{HTML}{9EC0A3}
\DeclareRobustCommand{\hlberry}[1]{{\sethlcolor{berry}\hl{#1}}}
\DeclareRobustCommand{\hlsage}[1]{{\sethlcolor{sage}\hl{#1}}}

\usepackage{subfig}

\title{Prompting Language Models for Linguistic Structure}

\author{Terra Blevins \quad Hila Gonen \quad Luke Zettlemoyer \\
        Paul G. Allen School of Computer Science \& Engineering,\\ University of Washington \\
        {\tt \{blvns, lsz\}@cs.washington.edu}\\
        {\tt hilagnn@gmail.com}}

\begin{document}
\maketitle
\begin{abstract}
Although pretrained language models (PLMs) can be prompted to perform a wide range of language tasks, it remains an open question how much this ability comes from generalizable linguistic understanding versus surface-level lexical patterns. To test this, we present a \textit{structured prompting} approach for linguistic structured prediction tasks, allowing us to perform zero- and few-shot sequence tagging with autoregressive PLMs. We evaluate this approach on part-of-speech tagging, named entity recognition, and sentence chunking, demonstrating strong few-shot performance in all cases. We also find that while PLMs contain significant prior knowledge of task labels due to task leakage into the pretraining corpus, structured prompting can also retrieve linguistic structure with arbitrary labels. These findings indicate that the in-context learning ability and linguistic knowledge of PLMs generalizes beyond memorization of their training data.
\end{abstract}

\section{Introduction}
The rapid increase in the scale of pretrained language models (PLMs) has led to a new paradigm of NLP modeling: in-context learning, or prompting \cite[e.g.,][]{brown2020language, raffel2020exploring}. In this setting, the model is used to perform a task directly via the predictions of the LM head without additional finetuning on the target task, often with a few demonstrations of the desired behavior provided within the input. This setup has led to impressive few-shot performance on various tasks ranging from classification to summarization and generation \cite{liu2021pre}.

Due to their broad success on tasks requiring language understanding, we hypothesize that these models also contain significant linguistic knowledge. However, we are not aware of existing prompting methods that can directly test this hypothesis on autoregressive PLMs. 
Behavioral analysis of PLMs \cite{belinkov2020interpretability} uses methods similar to prompting to measure knowledge stored in language models~\cite{gulordava2018colorless, petroni2019language}, but this technique is difficult to generalize to tasks that predict more complex structures. Additionally, current approaches for applying PLMs to linguistic structured prediction tasks finetune on the downstream task \cite[e.g.,][]{ma2021template}, which confounds measuring underlying model knowledge. 

%Structured Prompting Schematic 
\begin{figure}
    \centering
    \includegraphics[width=\linewidth]{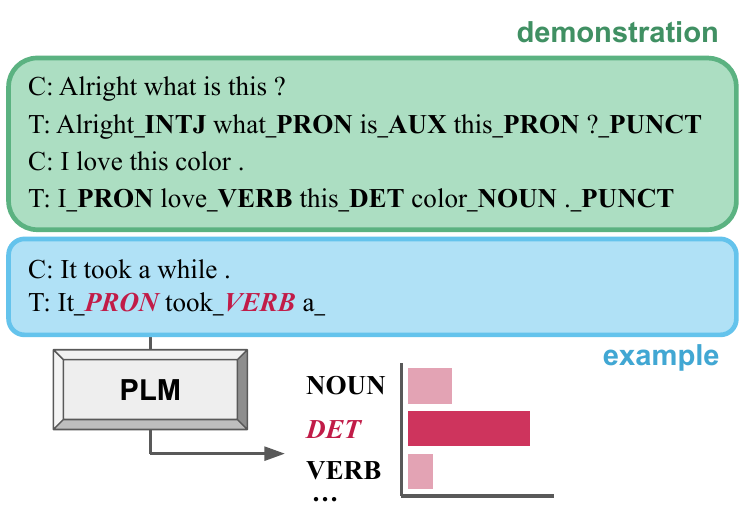}
    \caption{Sequence tagging via structured prompting. Each predicted label is appended to the context along with the next word to iteratively tag the full sentence.}
    \label{fig:prompt-examples}
\end{figure}

We propose a new approach, \textit{structured prompting}, that iteratively prompts autoregressive PLMs to probe for word- and span-level linguistics framed as sequence tagging tasks (Section \ref{sec:struct-prompting}). At timestep $t$, a label for the $t$-th word in the sequence is decoded from the LM; the model prediction is then fed back into the model along with the next word to progress to timestep $t+1$. We evaluate our approach on three sequence tagging tasks: POS tagging, sentence chunking, and NER. Our experiments show that PLMs can perform effective few-shot sequence tagging in the structured prompting setup, and that performance increases with the demonstration set size and model size, consistent with other prompting methods (Section \ref{sec:results}).

We further analyze structured prompting by examining how the model generalizes to various representations for labels (Section \ref{sec:analysis}) as well as by analyzing the presence of task data in the pretraining corpus and how this affects model performance (Section \ref{sec:data-analysis}). These experiments show that structured prompting can recover linguistic information from the model without using standard task labels, indicating that PLMs contain this knowledge in a general manner beyond memorization of the task from pretraining data. Interestingly, while PLMs perform best with meaningful labels (such as original task labels or full class names in English), the model can also in-context learn from arbitrary labels. Additionally, the model exhibits strong prior knowledge of the task labels' mapping onto the underlying classes, likely due to the prevalence of task data in the pretraining corpus.

The contributions of this work are therefore threefold: (1) we introduce a new paradigm, \textit{structured prompting}, that probes PLMs for sequence knowledge without further training, (2) we find that this approach recovers linguistic structure from PLMs in a few-shot manner, and (3) we present an analysis to quantify the effect of label form and pretraining data on in-context learning performance.
Overall, our findings provide insight into both the linguistic generalizations learned by PLMs and how in-context learning works in general.

\section{Structured Prompting of Pretrained Language Models}
\label{sec:struct-prompting}
We propose a sequential method for performing sequence tagging with PLMs via in-context learning, which we refer to as \textit{structured prompting} (Figure \ref{fig:prompt-examples}). 
The model is given $k$ (context, tagged sequence) pairs as the task demonstration and the example sentence to be labeled. The model then iteratively tags the words in the example with constrained decoding over a fixed set of labels.

More specifically, given a set of labels $L$ and an input sequence $c$ containing $k$ demonstration pairs as well as the full text of the example sentence $S = s_0, ..., s_n$, at each time step $t$ the language model $M$ encodes [$c; s_t$] and labels $s_t$ with $\hat{\ell}_t = \argmax\limits_{\ell \in L} P_M(\ell|c, s_t)$. We then update the input sequence by appending the current word $s_t$ and the predicted label $\hat{\ell}_t$ to the end of $c$. Multi-token labels are scored with the average log-likelihood over all tokens $P_M(\ell|c) = \frac{1}{|\ell|} \sum_{i=0}^{|\ell|} P_M(y_i|c,y_0,...,y_{i-1})$, where $y_j$ is the $j$th subword token in $\ell$. 

This approach to in-context learning tags an entire sequence with a single pass over the context. It also allows the model to condition on past predictions while labeling the current word.
As we demonstrate in Section \ref{sec:results}, these features allow us to apply large autoregressive language models to a broad class of core NLP tasks in a few-shot manner.
%(using model state caching and our custom decoding setup)

%Overall Results Fig for Section 4.1
\begin{figure*}
    \begin{center}
        \subfloat[]{\includegraphics[height=23ex]{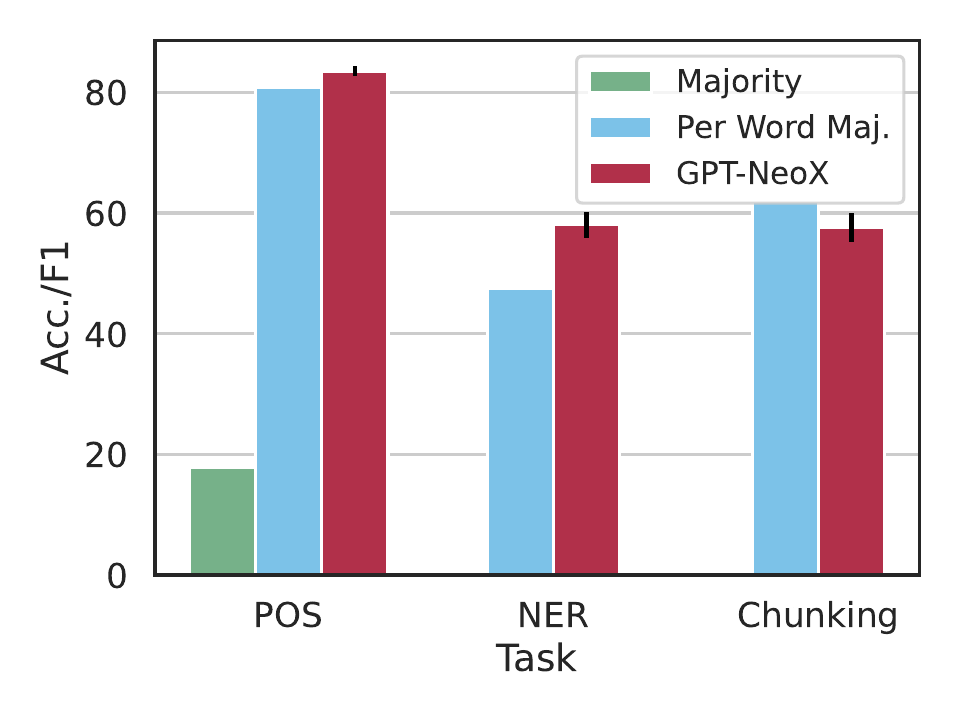}}
        \subfloat[]{\includegraphics[height=23ex]{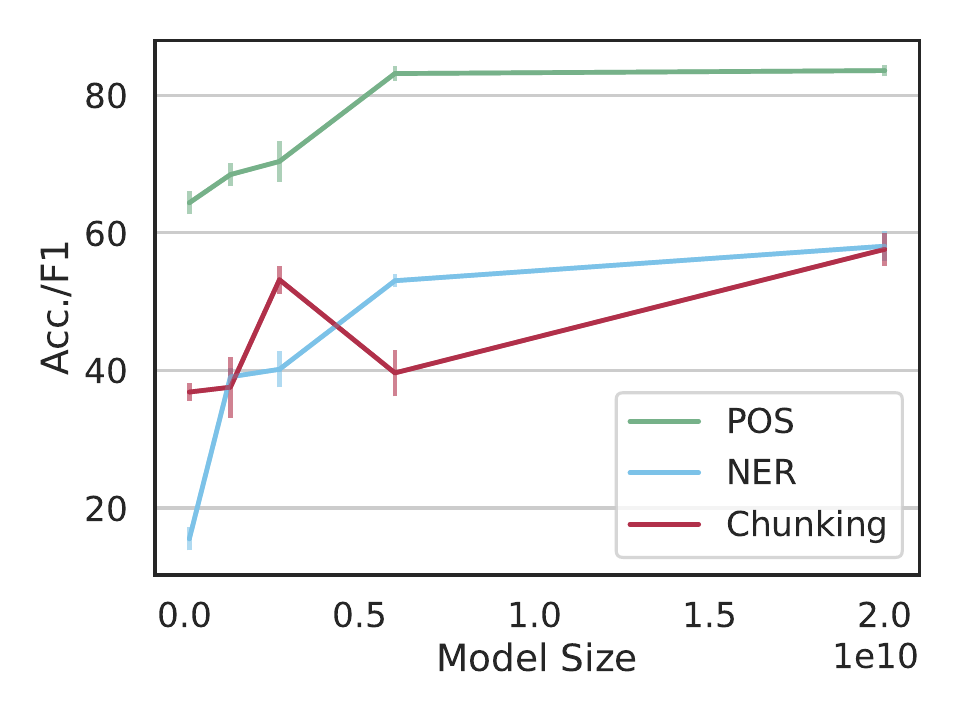}}
        \subfloat[]{\includegraphics[height=23ex]{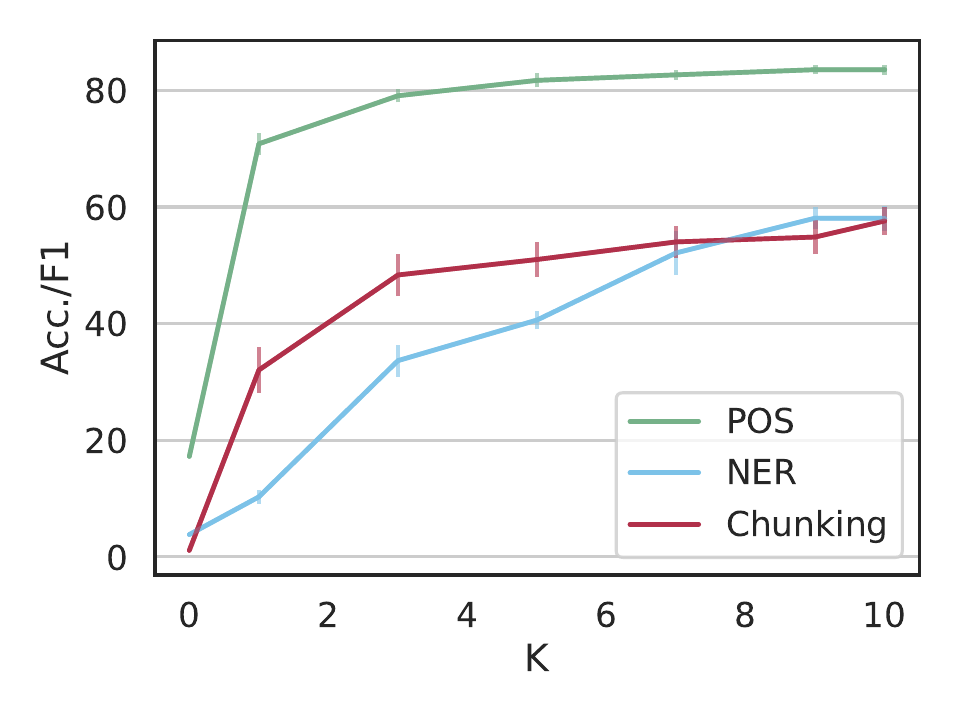}}
    \end{center}
    \caption{Results of the structured prompting evaluation. POS is evaluated on accuracy; the other tasks are evaluated with F1. (a) Results of GPT-NeoX (20B parameters) compared to task baselines. (b) Performance across different model sizes. (c) Performance of GPT-NeoX across different quantities of $k$ demonstrations.}
    \label{fig:overall_results}
\end{figure*}

\section{Experimental Setup}

\subsection{Prompt Formatting}
We use a lightweight prompt format with limited natural language guidance about the task provided to the model as shown in Figure \ref{fig:prompt-examples}; the letters ``C'' and ``T'' in the figure represent the inputs ``Context'' and ``Tagged'' respectively. For each task, we represent each tag with the token or sequence of tokens corresponding to the surface form of the label provided by the dataset.

In general, our preliminary experiments with varied prompt formats had little effect on performance. Specifically, performance was stable across the choice of delimiter and other minor formatting differences. However, we note that including the word in the ``Tagged'' sequence is important; on GPT-J, performance degrades by 84\% on POS and 79\% on NER when decoding the label sequence without repeating the word (i.e., ``Tagged: DET NOUN...'').
%(out of the considered options in $\{/, :, -, \_, =\}$)

\subsection{Sequence Tagging Tasks}
We consider the following English tasks framed as sequence tagging problems in evaluating the proposed structured prompting method. For tasks involving tagging spans of text, we label each token in the span using the \textit{BIO label format}: given a span of \textit{m} tokens labeled $\ell$, the first token is labeled as the beginning of the span with ``B-$\ell$'', the remaining \textit{m}-1 tokens are labeled as inside the span with ``I-$\ell$'', and tokens not included in the span are labeled as outside the span or ``O'').

\paragraph{Part-of-Speech (POS) Tagging} We evaluate POS tagging performance on English Universal Dependencies (UD) with the UPOS tagset \cite{nivre2020universal}. Specifically, we use the treebank annotated on the GUM corpus \cite{zeldes2017gum}.

\paragraph{Sentence Chunking} Chunking, or shallow parsing, partitions the words in a sentence into non-overlapping spans of syntactic meaning. We evaluate PLMs on chunking with the CONLL2000 dataset from \citet{sang2000introduction}, which frames chunking as a BIO tagging task. 

\paragraph{Named Entity Recognition (NER)} We evaluate the ability of structured prompting to extract named entities from PLMs with NER. This is measured as a BIO tagging task on the CONLL2003 dataset \cite{sang2003introduction}.

\subsection{Models}
We report performance on seven language models, ranging from 125 million to 175 billion parameters.

\paragraph{GPT-Neo} This set of PLMs contains models trained on the Pile \cite{gao2020pile} that from 125 million to 2.7 billion parameters \cite{gao2020pile}, 6.7 billion parameters \cite{wang2021gpt-j}, and 20 billion parameters \cite{black2022gpt-neox-20b}. We use the GPT-Neo models available through Huggingface \cite{wolf2019huggingface}.

\paragraph{GPT-3} We also perform structured prompting with the GPT-3 models \cite{brown2020language} via the OpenAI API. We use the base GPT-Curie ($\sim$6B parameters) and GPT-Davinci ($\sim$175B parameters) models that have undergone no additional instruction finetuning on POS tagging. Due to the cost of running these models through the API, we generate the GPT-Davinci output with unconstrained top-1 sampling rather than the constrained decoding setup described in Section \ref{sec:struct-prompting}.\newline

\noindent In preliminary experiments, we also tested structured prompting on several OPT models \cite{zhang2022opt}. We found their performance was significantly worse and did not scale with model size (up to 66B parameters) on POS tagging and NER. We leave a more thorough examination of this behavior discrepancy for future work.

\subsection{Additional Experimental Details}
We report the mean and standard error across $m$ runs for each experiment. For each of these runs, $k$ demonstrations are sampled from the training dataset at random, with the condition that the $k$ demonstrations cover the label space of the task if possible. We use $k=10$ sentences as demonstrations and perform $m=5$ runs per experiment unless otherwise stated.

Each model is evaluated on 1000 examples randomly sampled from the task test set (see Appendix \ref{app:analysis-eval} for a discussion on how this choice affects performance estimates). The evaluation subset is held fixed across all five runs, and the evaluation data and selection of demonstrations for each run are fixed across models for each task.

To obtain the tag sequence for each example, we greedily take the top-1 label (with the highest log likelihood) for each word. We also enforce hard constraints for the span-labeling tasks involving BIO tagging (chunking, NER) to ensure a valid BIO tag sequence (e.g., I-X tags can only follow a previous B-X or I-X tag). Empirically, we find that enforcing BIO constraints makes little difference in the method's overall performance; however, we use them as they ensure valid output sequences. Appendix \ref{app:analysis-bio} compares model performance with and without BIO constraints.

%Error Analysis for Sec. 4.2
\begin{figure*}
    \begin{center}
        \subfloat[]{\includegraphics[height=23ex]{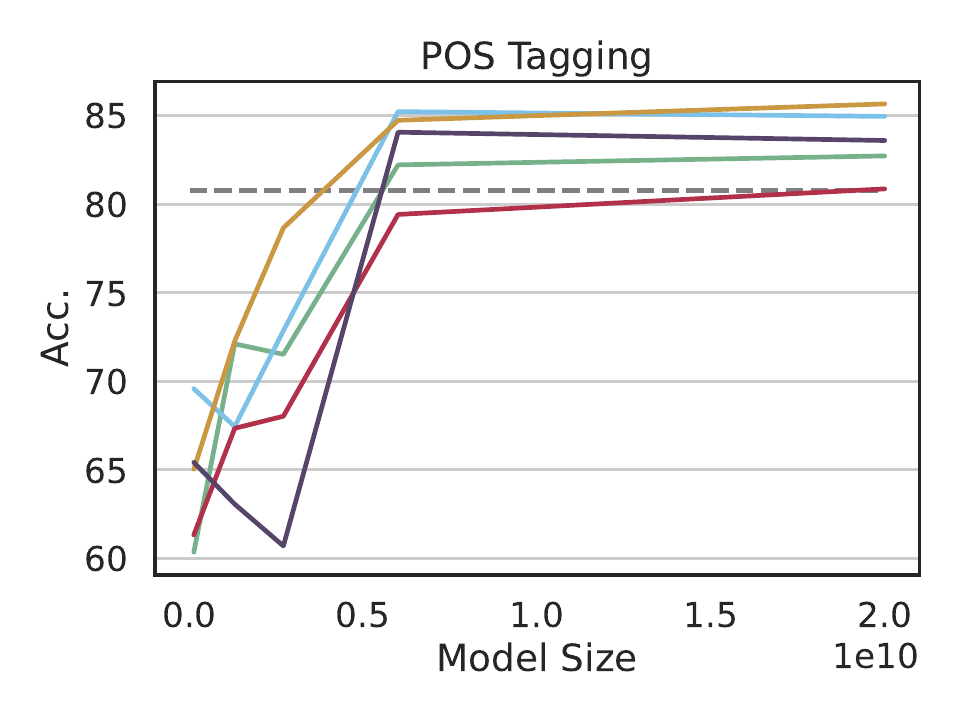}}
        \subfloat[]{\includegraphics[height=23ex]{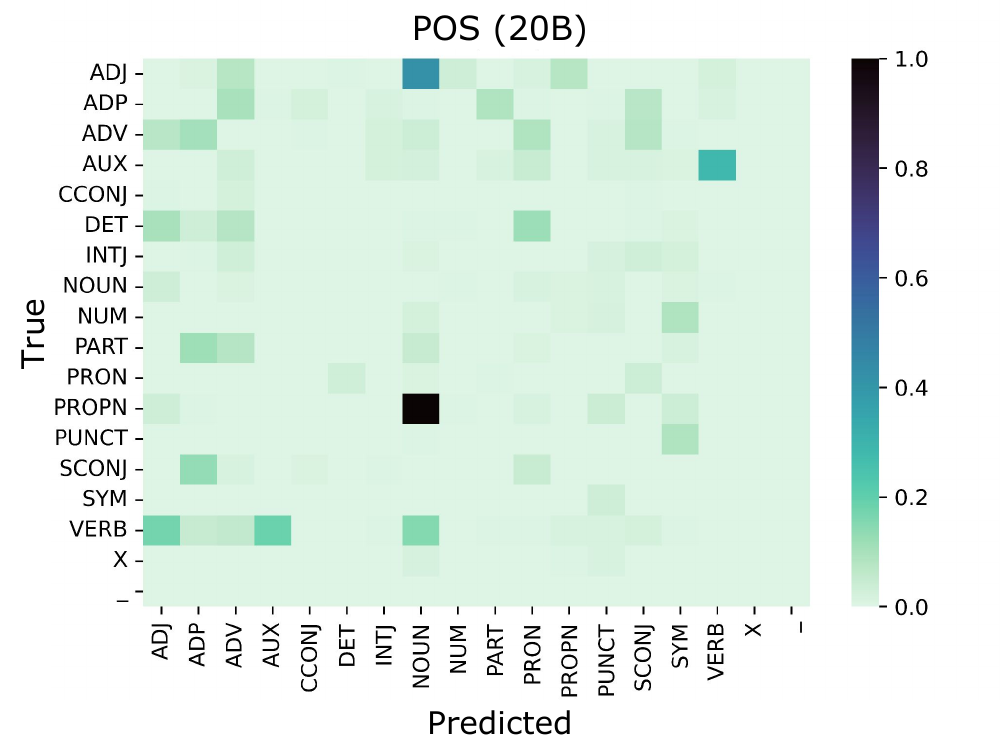}}
        \subfloat[]{\includegraphics[height=23ex]{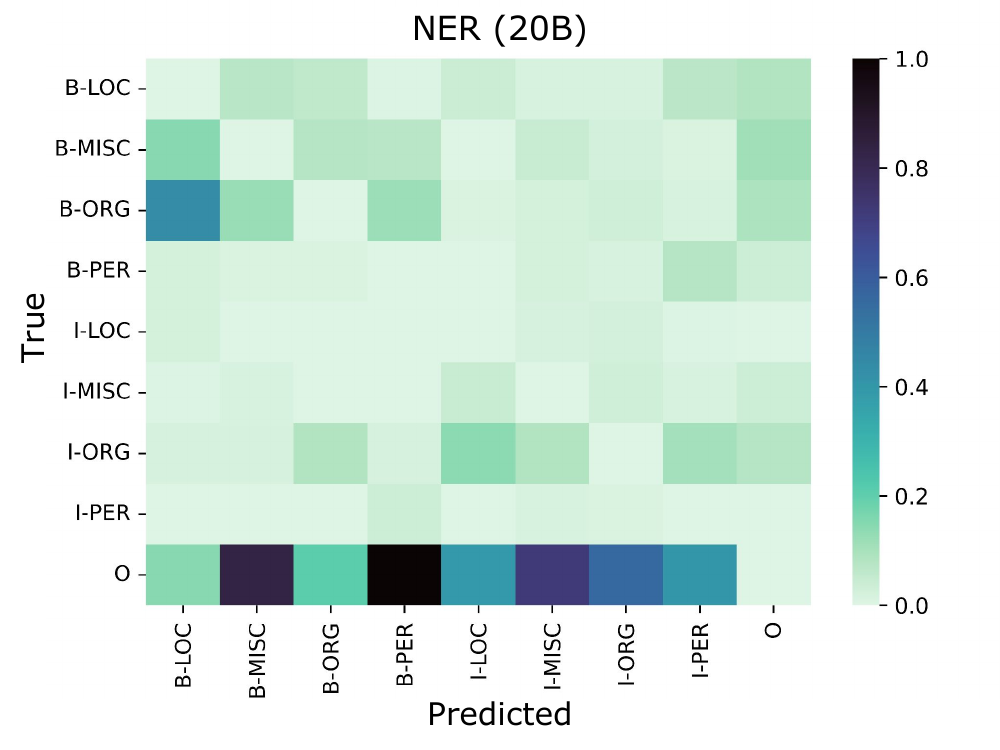}}
    \end{center}
    \caption{Error analysis of structured prompting for GPT-Neo series. (a) POS performance on different sets of 10-shot demonstrations. (b) Confusion matrix for GPT-NeoX on NER and (c) POS tagging, aggregated across runs.}
    \label{fig:error-analysis}
\end{figure*}

\section{Structured Prompting Results}
\label{sec:results}
We measure the performance of \textit{structured prompting} on three sequence tagging tasks. This evaluation aims to (1) validate that structured prompting follows prior prompting setups in terms of model and k-shot scaling trends and (2) investigate the extent to which the approach extracts these structures from the model. We then quantify the types of errors made with structured prompting. 

\subsection{Overall Results}
\label{sec:overall-results}
Figure \ref{fig:overall_results} presents the results of our primary structured prompting evaluation. We consider the performance of GPT-NeoX \cite{black2022gpt-neox-20b} compared to task baselines: \textit{overall majority}, in which each word is labeled with the most frequent tag in the training set, and \textit{per-word majority}, where each word is labeled with the tag it most commonly appeared within the training data (left panel).\footnote{For BIO tasks, the majority labels correspond to ``O'' (NER) and ``I-NP'' (chunking). The CONLL evaluation script only scores labeled spans, giving an overall majority F1 of 0.} All baselines are calculated on the full training set and so use more labeled data than the PLM; the per-word majority is a particularly strong baseline as words frequently occur with the same tag.  

\paragraph{Structured prompting performs effective few-shot sequence tagging} We find that GPT-NeoX significantly outperforms each baseline on POS tagging and NER, and the model slightly underperforms the per-word majority baseline on sentence chunking by 4.2 points. Overall, the approach performs worse for the BIO span-labeling tasks than for word-level POS tagging. We hypothesize that the former tasks are more complex, as they require the model to determine spans and more detailed linguistic knowledge. 

\paragraph{Structured prompting scales with model and demonstration size} We observe that the performance of structured prompting improves with scale across GPT-Neo models (center panel). Model performance also improves with additional demonstrations (right panel); both of these trends are consistent with prior prompting results \cite[e.g.,][]{black2022gpt-neox-20b}. However, the extent to which additional demonstrations help varies: NER improves more with larger sizes of $k$ than POS and chunking, likely because labeled spans are more sparse in NER. Notably, in the zero-shot case the model achieves around 17\% accuracy on POS tagging when randomly predicting labels would yield 5.8\%.

\begin{table}[]
    \centering
    \begin{tabular}{c c c|r r}
        \toprule
        \textbf{Size} & \textbf{Model} & \textbf{k} & \textbf{Acc.} & \textbf{SE}  \\
        \hline
        \multirow{2}{*}{$\sim$6B} & GPT-J$^{*}$ & 5 & 79.01 & 2.95 \\
        & GPT-Curie & 5 & 66.27 & 0.46 \\
        \hline
        \multirow{2}{*}{$\sim$175B} & GPT-Davinci$^{\dagger}$ & 5 & 59.65 & 2.84 \\
        & GPT-Davinci$^{\dagger}$ & 10 & 65.90 & 1.34 \\
        \toprule
    \end{tabular}
    \caption{Structured Prompting results on POS tagging for GPT-Curie and GPT-Davinci. SE is standard error. $^{*}$: model from GPT-Neo series of a similar size to Curie; $^{\dagger}$: evaluated with greedy unconstrained decoding.}
    \label{tab:gpt3-results}
\end{table}

\paragraph{Structured prompting with GPT-3}
Table \ref{tab:gpt3-results} compares two GPT-3 models to the GPT-Neo series on POS tagging.\footnote{Each experiment reported in this section is repeated across three runs rather than five.} We first compare the 6B parameter GPT-Curie \cite{gao2021sizes} to the similarly sized GPT-J model in a 5-shot setting. We find that GPT-Curie underperforms GPT-J by 12.7 points; both models also underperform the per-word majority baseline in this setting.

We then evaluate the largest GPT-3 model, GPT-Davinci, on POS tagging with greedy unconstrained decoding of the entire output sequence. Davinci performs reasonably well and scores similarly to Curie despite the more difficult decoding setting; many errors arise from format errors in the generated output for longer sentences. If we only evaluate examples that occur prior to these format errors, performance on that subset of the evaluation data is 72.85 ± 1.3 at k=5 and 78.04 ± 0.8 at k=10.

\subsection{Error Analysis}
\label{sec:error-analysis}
Figure \ref{fig:error-analysis} presents an error analysis of structured prompting; complete analyses for other tasks are provided in Appendix \ref{app:error-analysis}. We first break out performance across runs and evaluate how the choice of in-context examples affects performance (left panel). For POS tagging, the choice of demonstrations makes a difference, with some sets performing better than others across models and a performance gap of 4.8 accuracy points between the best and worst run on the 20B parameter model. NER exhibits similar results to POS; however, chunking performance of different demonstration sets is much more varied and inconsistent across models. 

Next, we examine common error types in structured prompting with confusion matrices (center and right panel). We zero out the diagonal (representing correct predictions) and normalize the matrices for clarity. Many of the mistakes made by the 20B parameter model on POS tagging are for syntactically similar roles, such as confusing proper nouns for nouns and labeling auxiliary verbs as verbs. However, for BIO tagging the models are not always well-calibrated: on NER, the model most often mislabels ``O'' tokens, indicating that the model overpredicts named entities.

Given that the choice of demonstrations affects PLM performance, another consideration is: how consistent are the error types across runs? To investigate this, we calculate the pairwise Spearman correlations between the confusion matrices of each run. These correlations are very high for the 20B parameter model, indicating the model makes similar types of error across runs: on average $\rho = 0.77$ for POS tagging, 0.83 for NER, and 0.88 for chunking; all pairwise correlations have p-values $<<0.001$. Additionally, the models seem to become more robust across demonstration sets at scale; confusion matrix correlations for the 2.7B model are lower ($\rho = 0.71, 0.64, 0.66$ for POS, NER, and chunking, respectively).

\section{When Does Structured Prompting Work?}
\label{sec:analysis}
We now investigate how structured prompting surfaces linguistic structure from PLMs, using the behavior of GPT-NeoX on POS tagging and NER as a case study. 
We find that (1) in some cases, the model generalizes to labels not seen in the demonstration, and (2) the label form has a large effect on performance.
Specifically, the model can learn in context when arbitrary labels represent classes but will ignore label mappings in the demonstration that contradict its prior task knowledge.

%Seen vs. unseen labels
\begin{figure}
    \centering
    \includegraphics[width=0.7\linewidth]{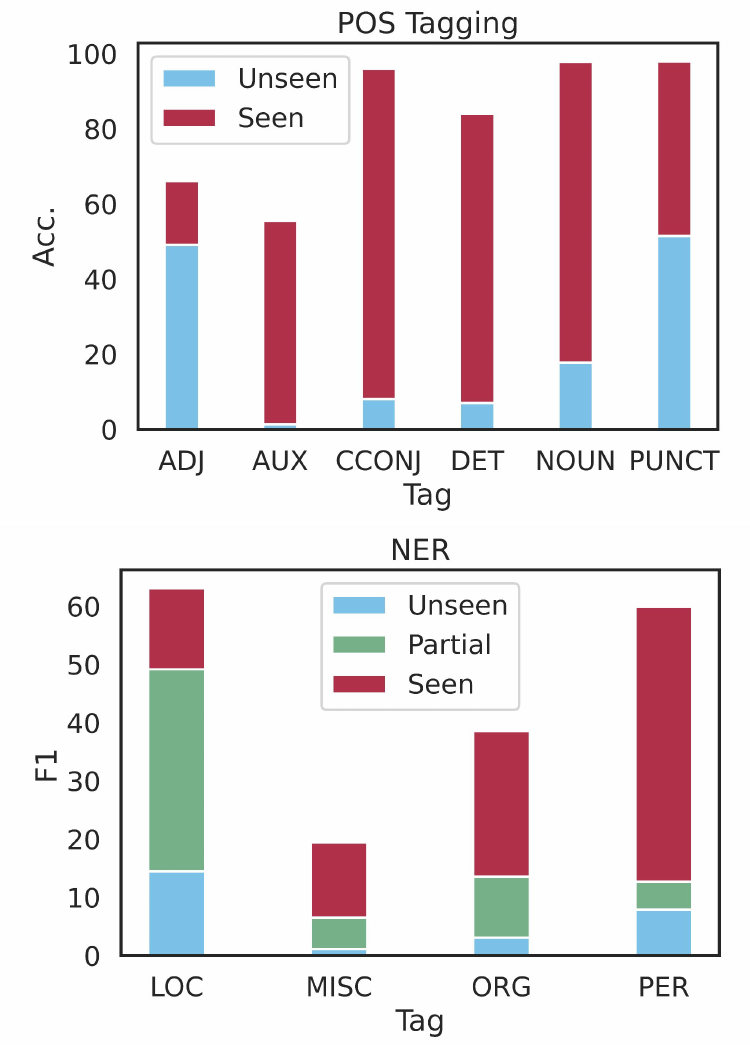}
    \caption{Performance of GPT-NeoX when the labels are seen vs. not seen in the demonstration. \textit{Partial} labels are those only seen as B-<label> tags in BIO tagging.}
    \label{fig:seen-labels}
\end{figure}

\subsection{Effect of Seen Labels}
\label{sec:analysis-seen-labels}
In Section \ref{sec:overall-results}, we see that the model obtains above random chance accuracy on zero-shot POS tagging, suggesting that the model does not need to observe the label to associate it with the correct class. To analyze this, we compare the model's performance when the label is and is not seen in the demonstration, averaged across k-shot runs.

Model performance on unseen tags, and the gain in performance after observing the tag, varies greatly by label class (Figure \ref{fig:seen-labels}). For some classes in POS tagging, such as ADJ and PUNCT, the model obtains around 50\% accuracy without seeing the label. However, unseen performance on AUX in POS tagging and MISC in NER is close to 0\%. Furthermore, while observing tags like LOC in NER greatly improves performance, other tags like ADJ and MISC improve much less when seen.

%Label Ablations Figure
\begin{figure}
    \centering
    \includegraphics[width=0.9\linewidth]{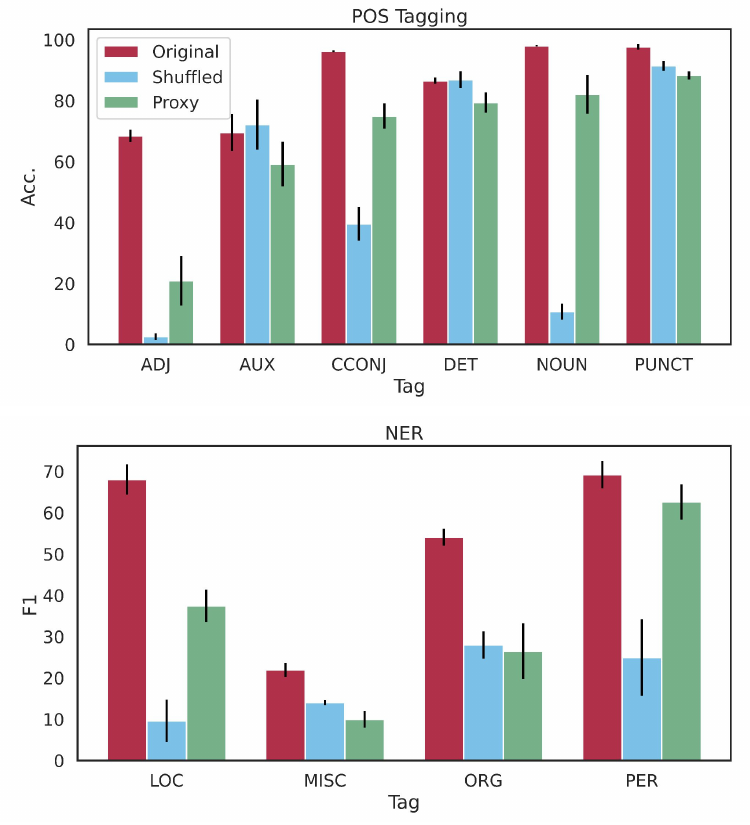}
    \caption{Results of ablating the surface form of the labels for structured prompting. }
    \label{fig:label-ablation}
\end{figure}

\subsection{Effect of Label Form}
\label{sec:analysis-label-effect}
We hypothesize that the behavior observed in Section \ref{sec:analysis-seen-labels} depends on how informative the label form is for the class. Therefore, we compare the model performance on (1) the \textit{original} task labels; (2) \textit{shuffled} task labels, where we shuffle the label surface forms but maintain underlying class correspondences to words; and (3) \textit{proxy} labels, where we represent the classes with arbitrary tokens -- here, consecutive integers ranging from 11 to 27 (POS) and from 11 to 14 (NER). (Figure \ref{fig:label-ablation}).

\paragraph{Label shuffling confuses GPT-NeoX} Shuffling the labels greatly hurts overall model performance, with POS scores decreasing overall by 50.5\%, and NER by 65.9\%. Some classes are more robust to the shuffled labels than others: the AUX and DET parts-of-speech score within the standard error of the original class performance, whereas ADJ accuracy drops by 96.2\% to near zero.

Interestingly, most mistakes made in the \textit{shuffled} setting (61.4\%) result from the model predicting the true class label rather than the shuffled one from the demonstration. This occurs more frequently for classes whose performance severely degrades when shuffled: 93.9\% of errors on the NOUN class are due to this phenomenon, and across classes, there is a strong correlation between performance degradation and the percent of errors predicting the true label ($\rho=0.69, p<0.05$). 
This result suggests that PLMs ignore in-context label mappings when the model already associates the label with a specific class, similar to findings in \citet{min2022rethinking}.

\paragraph{GPT-NeoX in-context learns with arbitrary proxy labels} Model behavior with the proxy labels is closer to the original labels, with performance decreasing by 25.8\% on POS and 30.5\% on NER. Indeed, on many labels that significantly degrade with label shuffling, the model performs significantly better on the proxy labels (NOUN and CCONJ in POS tagging, PER in NER). These results demonstrate that the model is able to perform in-context learning to extract linguistic structure, even when the tags are uninformative.

%Data Analysis Results Table
\begin{table*}[]
    \centering
    \small
    \begin{tabular}{l|r r|l}
        \toprule
        \textbf{Label} & \textbf{Freq.} & \textbf{Task Stats} & \textbf{Example Contexts} \\
        \hline
        \hline
        \multicolumn{2}{l}{\textbf{POS Tagging}} & \textbf{UD Format} & \\
        \hline
        \rowcolor{grey}
        & & & The 10 most frequent relations where parent and child node agree in `Polarity`: \\
        \rowcolor{grey}
        \multirow{-2}{*}{NOUN} & \multirow{-2}{*}{360k} & \multirow{-2}{*}{9.29\%} &
        <tt>\textbf{NOUN} ----> ADJ</tt> (2; 100\%) (GitHub)\\
        CCONJ & 22k & 23.48\% & 13 \textbackslash t und \textbackslash t und \textbackslash t \textbf{CCONJ} \textbackslash t KON \textbackslash t \_ \textbackslash t 14 \textbackslash t cc (GitHub) \\
        \rowcolor{grey}
        DET & 1.53M & 0.72\% & \textbf{DET}: determiner, e.g. a, an, the \textbackslash n INTJ: interjection, e.g. psst... (StackExchange) \\
        \hline
        \hline
        \multicolumn{2}{l}{\textbf{NER}} & \textbf{Relevant?}
        \\
        \hline
        \rowcolor{grey}
         & & & Bacterial pellets were lysed in 10 ml \textbf{B-PER} Bacterial Protein Extraction \\
        \rowcolor{grey}
        \multirow{-2}{*}{B-PER} & \multirow{-2}{*}{5,655} & \multirow{-2}{*}{26/100} & Reagent... (PubMed)\\
        I-LOC & 2,197 & 43/100 & y = np.asarray("B-PER O O B-LOC \textbf{I-LOC} O B-ORG".split()) (StackExchange) \\
        \rowcolor{grey}
         & & & *I-PER* label usually follows *B-PER* and *I-PER*, but it cannot follow \\
        \rowcolor{grey}
        \multirow{-2}{*}{B-ORG} & \multirow{-2}{*}{2603} & \multirow{-2}{*}{80/100} & *\textbf{B-ORG}* or *I-ORG*. (Arxiv)\\
        
        I-MISC & 907 & 76/100 & My(O) favorite(O) book(O) is(O) harry(B-MISC) potter(\textbf{I-MISC})... (StackExchange) \\
        \toprule
    \end{tabular}
    \caption{Analysis of the Pile for labels from UD POS tagset and CONLL03 NER tagset. \textit{Task Stats} document the percentage of occurrences that are in the UD format for POS tagging and the proportion of sampled documents relevant to NER. Some examples are slightly edited for readability.}
    \label{tab:data-analysis}
\end{table*}

\section{Sources of Linguistic Knowledge in Pretraining Corpus}
\label{sec:data-analysis}
The results in Section \ref{sec:analysis} demonstrate that the choice of label form can greatly affect structured prompting performance and implies that the model contains prior task knowledge. We analyze contexts in which the labels for POS tagging and NER appear in the Pile \cite{gao2020pile} to better understand what, if any, task information GPT-NeoX learns from pretraining.

Our analysis shows that task information occurs in the pretraining data, both as labeled examples (Section \ref{sec:data-contamination}) and in other related contexts (Section \ref{sec:data-other}). However, we find no evidence of test data leakage. Given these findings, we evaluate the model in a new setting that substitutes an English description of each class (e.g., ``adjective'', ``person'') for the label in order to control for label leakage while still providing meaningful labels (Section \ref{sec:data-relationship}). 

\subsection{Task Data Contamination}
\label{sec:data-contamination}

A likely location for task labels to occur is leaked task examples from pretraining data sources. To test this, we search the Pile for instances of labeled POS and NER data (Table \ref{tab:data-analysis}, the full results are given in Appendix \ref{app:data-analysis}). 

\paragraph{POS Tagging} Since the POS data is obtained from UD treebanks, we search the Pile for each label as it would appear in the treebank (with tab whitespace on either side of it, see CCONJ example context). We find a significant amount of UD data formatted in this manner: up to 33,000 occurrences for an individual label (NOUN). This is unsurprising given that Github -- where UD treebanks are hosted -- is a data source for the Pile. However, we find no evidence of test data leakage across any of the POS label occurrences when compared to the GUM treebank \cite{zeldes2017gum}.\footnote{We also compare the test set against the Pile via other methods (exact document match and searching for individual lines); none of these match any test data against the Pile.} 

We also perform a closer analysis of the CCONJ label: we compare each occurrence against all nine English treebanks in UD and manually examine it.
We find that many CCONJ occurrences can be found in the English Web Treebank \cite[EWT;][]{silveira14gold} (1052/118/155 from the train/dev/test splits); others match with Parallel Universal Dependencies \cite[PUD;][]{zeman2017conll} (10 occurrences from test set) and ParaTUT \cite{sanguinetti2014converting} (1 occurrence from development set). 

Our manual analysis finds that most of the CCONJ occurrences are in non-English documents (77\%); other languages whose treebanks we see include Finnish, German, and Arabic, among many others.\footnote{This is unsurprising: though the Pile is characterized as an "English text corpus" \cite{gao2020pile}, prior work has found similar corpora derived from the web contain significant amounts of non-English text \cite{blevins2022language}.} We also observe that every tab-separated instance of CCONJ occurs in the UD treebank format, indicating that this automatic filter is a reasonable estimate of UD data leakage across labels.

\paragraph{NER} Task data leakage for NER is much more limited than POS: the most frequent label occurs 5,655 times in the Pile (other than ``O'' which occurs very frequently in many contexts). Since the CONLL format separates the tags with spaces instead of tabs, it is more difficult to filter for data leakage. Instead, we manually evaluate 100 examples for the BIO labels and give the proportion of the sample that is relevant for NER. 

Only a subset of relevant occurrences includes labeled data -- our analysis found that labeled data is not common, and most cases are single example sentences annotated in various ways that do not necessarily follow the CONLL format (see I-MISC example context). Similar to POS tagging, we also find labeled examples in non-English languages; notably, some of the examples observed are incorrectly labeled.\footnote{For example, the phrase ``l'entreprise/O SpaceX/O...'' occurs in a WebText2 document; however, SpaceX is a named entity that should be labeled as B-ORG.} This highlights that while the model sees task data during pretraining, the quality and accuracy of that data are unverified.

\subsection{Labels in Other Contexts}
\label{sec:data-other}

During the data analysis, we also observe tags from our tasks in settings other than labeled data. Other relevant contexts are task documentation or descriptions (see NOUN, DET, and B-ORG example contexts) and code related to the task (I-LOC example context). These contexts are particularly interesting, as they provide information that may help the model learn by explaining the task in natural language or code, rather than via input/output pairs.

We also observe instances of labels that are unrelated to the task. This is more common for the POS tags; whereas, for NER labels, up to 80\% of the sampled contexts are related to the task. The topic of these unrelated contexts varies widely across labels, from biomedical and legal texts (see B-PER example context) to unrelated source code and news articles. 

%Label Correlations Table
\begin{table}[]
    \centering
    \small
    \begin{tabular}{c c | r r r r }
        \toprule
        &  & \multicolumn{4}{c}{\textbf{Label Sets}} \\
        &  & Origin. & Shuffle & Proxy & Words \\
        \hline
        \hline
        \multicolumn{5}{l}{\textbf{POS Tagging}}\\
        \hline
        \multirow{4}{*}{\makecell{$\Delta$\\Acc.}} & Origin. & \cellcolor{sage} 83.55 & \cellcolor{grey}  & \cellcolor{grey}  & \cellcolor{grey}  \\
        & Shuffle & -42.11 & \cellcolor{sage} 41.44 & \cellcolor{grey} & \cellcolor{grey} \\
        & Proxy & -21.57 & 20.54 & \cellcolor{sage} 61.98 & \cellcolor{grey} \\
        & Words & -5.43 & 36.67 & 16.13  & \cellcolor{sage} 78.11 \\
        \hline 
        \multirow{4}{*}{$\rho$} & Origin. & 1 & \cellcolor{grey} & \cellcolor{grey} & \cellcolor{grey} \\
        & Shuffle & 0.676 & 1 & \cellcolor{grey} & \cellcolor{grey} \\
        & Proxy & \cellcolor{berry} 0.934* & 0.718 & 1 & \cellcolor{grey} \\
        & Words & \cellcolor{berry} 0.924* & 0.667 & \cellcolor{berry} 0.909* & 1 \\
        \hline
        \hline
        \multicolumn{5}{l}{\textbf{NER}}\\
        \hline
         \multirow{4}{*}{\makecell{$\Delta$\\F1}} & Origin. & \cellcolor{sage} 58.05 & \cellcolor{grey} & \cellcolor{grey} & \cellcolor{grey} \\
        & Shuffle & -38.28 & \cellcolor{sage} 19.77 & \cellcolor{grey} & \cellcolor{grey} \\
        & Proxy & -17.65 & 20.63 & \cellcolor{sage} 40.40 & \cellcolor{grey} \\
        & Words & -1.17$^{\dagger}$ & 37.11 & -16.48 & \cellcolor{sage} 56.88 \\
        \toprule
    \end{tabular}
    \caption{Performance deltas ($\Delta$, column - row) and spearman correlations ($\rho$) of classes between label sets. \hlsage{$\Delta$ diagonals} report performance with that set. $\dagger$: delta is within standard error; \hlberry{*: p <{}< 0.001.}}
    \label{tab:label_corrs}
\end{table}

\subsection{Relationship Between Labels and Classes}
\label{sec:data-relationship}

Due to the quantity of task data uncovered in the Pile, we would like to control for the effect of pretraining on labeled data. To this end, we evaluate GPT-NeoX on semantically meaningful labels not previously seen in labeled contexts; specifically, we replace the task labels with the English name for each class (e.g., adjective, B-location), which we refer to as the \textit{words} label set. The model achieves an accuracy of 78.11 ± 1.46 on POS tagging and an F1 score of 56.88 ± 0.86 for NER in this setting. 

In Table \ref{tab:label_corrs}, we compare the performance between these label sets and evaluate how correlated individual class performances are across these sets. We observe an identical ranking across label sets in POS tagging and NER. On NER, the difference in model performance between the \textit{true} labels and \textit{words} as labels is within standard error. However, on POS there is a small but significant decrease of 5.4 points between the two; this drop in performance likely quantifies the benefit of observing the POS task data in the Pile. 

The correlation study shows that performance across classes on the \textit{original}, \textit{proxy}, and \textit{words} label sets for POS tagging are all strongly correlated ($\rho$ > 0.9). However, their correlations with the \textit{shuffled} labels are less significant; this difference is likely due to the prior task knowledge GPT-NeoX has for UD labels leading to predicting the actual label of the class rather than the shuffled one, as seen in Section \ref{sec:analysis-label-effect}.

\section{Related Work}
\paragraph{Prompting PLMs for Sequence Information}
Recent work has applied various prompting approaches to sequence tagging tasks, primarily focusing on NER \cite{cui2021template, ma2021template}. However, these approaches also require further training, most often by learning new prompt embeddings for the task \cite[][]{li2022probing, liu2022p, chen2022lightner}. Other work has finetuned language models to apply them to sequence tagging tasks \cite[][]{liu2022autoregressive}. In contrast, our approach requires no additional parameters to be learned. More similar to our work is the sequence tagging method in \citet{shliazhko2022mgpt}, though their approach prompts the model separately for each word in the sentence.
Additionally, similar approaches to prompting have been proposed for other tasks; these methods decompose a target task and repeatedly prompt the model on subtasks, building on the model's outputs to generate the final prediction \cite{zhou2022least, press2022measuring}. 
However, these approaches solve a different subset of NLP tasks and use the outputs from the intermediate prompting steps differently (i.e., by conditioning on them in future prompting steps, whereas in structured prompting each output is a predicted label).

\paragraph{Probing Pretrained Models}
There is extensive work on probing models for their underlying knowledge \cite[inter alia.]{belinkov2017evaluating, blevins2018deep, gulordava2018colorless}. The approach has become particularly popular for analyzing masked PLMs \cite[e.g.,][]{liu2019linguistic, liu2021probing}, with behavioral probes \cite[e.g.][]{petroni2019language, balasubramanian2020whats} in particular using the LM setup to elicit knowledge from the model.

However, prompting autoregressive PLMs \cite{brown2020language, schick2021its, gao2021making}, though technically similar to behavioral probing, is usually not framed as probing the underlying model for knowledge. Some exceptions are \citet{alivanistos2022prompting}, which uses prompting techniques to probe the LM for knowledge base relations, and \citet{li2022probing}, which replaces diagnostic probes with trained prompt embeddings for model analysis. We extend this framing by applying structured prompting as a behavioral probe for linguistic structure.

\paragraph{Analysis of Prompting Methods}
The results of the structured prompting setup ablations are consistent with prior work. Specifically, our observation of the model's prior label knowledge is similar to \citet{min2022rethinking}. We expand on their findings by showing that the model can still perform in-context learning with proxy labels where the model has no prior mapping for the task.

Other work has also documented the presence of task data in common pretraining corpora \cite{dodge2021documenting}, shown the effect of pretraining term frequencies on in-context performance \cite{razeghi2022impact}, and demonstrated the ability of LMs to learn from task data during pretraining \cite{magar2022data}. Similarly, we document the presence of task data and labels in the Pile and find that this signal can help task performance due to the model prior over the labels.

\section{Conclusion}
We propose \textit{structured prompting}, a general paradigm for sequence tagging with autoregressive PLMs. Our experiments show structured prompting performs well on three few-shot sequence tagging tasks. Further analysis shows that (1) the approach can elicit linguistic structure in many settings, including when the labels are unrelated to the task, and (2) while labeled task data is present in the pretraining corpora, using informative labels not found in task data gives similar performance to using the task labels. These findings indicate that the model's knowledge of linguistic structure is more general than the memorization of the task data. More generally, our approach provides a method to probe PLMs for sequence knowledge without training new or existing parameters.

\section*{Limitatons}
\paragraph{Data Leakage}
As discussed in Section \ref{sec:data-contamination}, we find evidence of labeled task data for POS tagging and (to a more limited extent) NER in the Pile. We attempt to control for this leakage by evaluating with class names as labels rather than the original tag set; however, due to the cost of training recent PLMs and their large pretraining corpora, it is impossible to control for data leakage when prompting existing models completely.

Both \citet{brown2020language} and \citet{chowdhery2022palm} discuss the presence of task data in their pretraining corpora when training PLMs and the difficulty of controlling for it in their evaluations. For downstream users, this issue is further compounded in cases where the pretraining data is unavailable, as it is impossible to even check for contamination in those cases (such as our GPT-3 experiments).

\paragraph{Experimental Limitations with GPT-3}
We only perform a subset of our evaluations of structured prompting on GPT-3, due to the cost of running the models in the API; this also means we do not run comprehensive prompt ablations to better tailor the setup for these models. Additionally, the results (i.e., lower performance than comparable GPT-Neo models) are difficult to interpret due to the black box nature of the GPT-3 models -- it may be due to pretraining data differences (as mentioned in the previous limitation), the lack of prompt engineering for the models, or some other discrepancy.

\paragraph{English-only Experiments}
The experiments in this paper focus on English sequence tagging tasks, and it is unclear how well the proposed method generalizes to other languages. We find evidence of task-relevant data in pretraining corpora in non-English languages, which suggests there is signal for the approach to work in other languages. However, prior work shows that PLMs behave much worse when prompted outside of English \cite{lin2021few, shi2022language} but does not address the effect of pretraining data on this phenomenon.

\section*{Acknowledgements}
We would like to thank Sewon Min and Ari Holtzman for their helpful conversations about the work.

\bibliography{custom}

\begin{thebibliography}{44}
\expandafter\ifx\csname natexlab\endcsname\relax\def\natexlab#1{#1}\fi

\bibitem[{Alivanistos et~al.(2022)Alivanistos, Santamar{\'\i}a, Cochez, Kalo,
  van Krieken, and Thanapalasingam}]{alivanistos2022prompting}
Dimitrios Alivanistos, Selene~B{\'a}ez Santamar{\'\i}a, Michael Cochez,
  Jan-Christoph Kalo, Emile van Krieken, and Thiviyan Thanapalasingam. 2022.
\newblock Prompting as probing: Using language models for knowledge base
  construction.
\newblock In \emph{LM-KBC 22: Knowledge Base Construction from Pre-trained
  Language Models}.

\bibitem[{Balasubramanian et~al.(2020)Balasubramanian, Jain, Jindal, Awasthi,
  and Sarawagi}]{balasubramanian2020whats}
Sriram Balasubramanian, Naman Jain, Gaurav Jindal, Abhijeet Awasthi, and Sunita
  Sarawagi. 2020.
\newblock \href {https://doi.org/10.18653/v1/2020.repl4nlp-1.24} {What{'}s in a
  name? are {BERT} named entity representations just as good for any other
  name?}
\newblock In \emph{Proceedings of the 5th Workshop on Representation Learning
  for NLP}, pages 205--214, Online. Association for Computational Linguistics.

\bibitem[{Belinkov et~al.(2020)Belinkov, Gehrmann, and
  Pavlick}]{belinkov2020interpretability}
Yonatan Belinkov, Sebastian Gehrmann, and Ellie Pavlick. 2020.
\newblock \href {https://doi.org/10.18653/v1/2020.acl-tutorials.1}
  {Interpretability and analysis in neural {NLP}}.
\newblock In \emph{Proceedings of the 58th Annual Meeting of the Association
  for Computational Linguistics: Tutorial Abstracts}, pages 1--5, Online.
  Association for Computational Linguistics.

\bibitem[{Belinkov et~al.(2017)Belinkov, M{\`a}rquez, Sajjad, Durrani, Dalvi,
  and Glass}]{belinkov2017evaluating}
Yonatan Belinkov, Llu{\'\i}s M{\`a}rquez, Hassan Sajjad, Nadir Durrani, Fahim
  Dalvi, and James Glass. 2017.
\newblock Evaluating layers of representation in neural machine translation on
  part-of-speech and semantic tagging tasks.
\newblock In \emph{Proceedings of the Eighth International Joint Conference on
  Natural Language Processing (Volume 1: Long Papers)}, pages 1--10.

\bibitem[{Black et~al.(2022)Black, Biderman, Hallahan, Anthony, Gao, Golding,
  He, Leahy, McDonell, Phang, Pieler, Prashanth, Purohit, Reynolds, Tow, Wang,
  and Weinbach}]{black2022gpt-neox-20b}
Sid Black, Stella Biderman, Eric Hallahan, Quentin Anthony, Leo Gao, Laurence
  Golding, Horace He, Connor Leahy, Kyle McDonell, Jason Phang, Michael Pieler,
  USVSN~Sai Prashanth, Shivanshu Purohit, Laria Reynolds, Jonathan Tow, Ben
  Wang, and Samuel Weinbach. 2022.
\newblock \href {https://arxiv.org/abs/2204.06745} {{GPT-NeoX-20B}: An
  open-source autoregressive language model}.
\newblock In \emph{Proceedings of the ACL Workshop on Challenges \&
  Perspectives in Creating Large Language Models}.

\bibitem[{Blevins et~al.(2018)Blevins, Levy, and Zettlemoyer}]{blevins2018deep}
Terra Blevins, Omer Levy, and Luke Zettlemoyer. 2018.
\newblock Deep {RNNs} encode soft hierarchical syntax.
\newblock In \emph{Proceedings of the 56th Annual Meeting of the Association
  for Computational Linguistics (Volume 2: Short Papers)}, pages 14--19.

\bibitem[{Blevins and Zettlemoyer(2022)}]{blevins2022language}
Terra Blevins and Luke Zettlemoyer. 2022.
\newblock Language contamination helps explain the cross-lingual capabilities
  of {E}nglish pretrained models.
\newblock In \emph{Proceedings of the 2022 Conference on Empirical Methods in
  Natural Language Processing (EMNLP)}.

\bibitem[{Brown et~al.(2020)Brown, Mann, Ryder, Subbiah, Kaplan, Dhariwal,
  Neelakantan, Shyam, Sastry, Askell et~al.}]{brown2020language}
Tom Brown, Benjamin Mann, Nick Ryder, Melanie Subbiah, Jared~D Kaplan, Prafulla
  Dhariwal, Arvind Neelakantan, Pranav Shyam, Girish Sastry, Amanda Askell,
  et~al. 2020.
\newblock Language models are few-shot learners.
\newblock \emph{Advances in neural information processing systems},
  33:1877--1901.

\bibitem[{Chen et~al.(2022)Chen, Li, Deng, Tan, Xu, Huang, Si, Chen, and
  Zhang}]{chen2022lightner}
Xiang Chen, Lei Li, Shumin Deng, Chuanqi Tan, Changliang Xu, Fei Huang, Luo Si,
  Huajun Chen, and Ningyu Zhang. 2022.
\newblock \href {https://aclanthology.org/2022.coling-1.209} {{L}ight{NER}: A
  lightweight tuning paradigm for low-resource {NER} via pluggable prompting}.
\newblock In \emph{Proceedings of the 29th International Conference on
  Computational Linguistics}, pages 2374--2387, Gyeongju, Republic of Korea.
  International Committee on Computational Linguistics.

\bibitem[{Chowdhery et~al.(2022)Chowdhery, Narang, Devlin, Bosma, Mishra,
  Roberts, Barham, Chung, Sutton, Gehrmann et~al.}]{chowdhery2022palm}
Aakanksha Chowdhery, Sharan Narang, Jacob Devlin, Maarten Bosma, Gaurav Mishra,
  Adam Roberts, Paul Barham, Hyung~Won Chung, Charles Sutton, Sebastian
  Gehrmann, et~al. 2022.
\newblock Palm: Scaling language modeling with pathways.
\newblock \emph{arXiv preprint arXiv:2204.02311}.

\bibitem[{Cui et~al.(2021)Cui, Wu, Liu, Yang, and Zhang}]{cui2021template}
Leyang Cui, Yu~Wu, Jian Liu, Sen Yang, and Yue Zhang. 2021.
\newblock \href {https://doi.org/10.18653/v1/2021.findings-acl.161}
  {Template-based named entity recognition using {BART}}.
\newblock In \emph{Findings of the Association for Computational Linguistics:
  ACL-IJCNLP 2021}, pages 1835--1845, Online. Association for Computational
  Linguistics.

\bibitem[{Dodge et~al.(2021)Dodge, Sap, Marasovi{\'c}, Agnew, Ilharco,
  Groeneveld, Mitchell, and Gardner}]{dodge2021documenting}
Jesse Dodge, Maarten Sap, Ana Marasovi{\'c}, William Agnew, Gabriel Ilharco,
  Dirk Groeneveld, Margaret Mitchell, and Matt Gardner. 2021.
\newblock Documenting large webtext corpora: A case study on the colossal clean
  crawled corpus.
\newblock In \emph{Proceedings of the 2021 Conference on Empirical Methods in
  Natural Language Processing}, pages 1286--1305.

\bibitem[{Gao(2021)}]{gao2021sizes}
Leo Gao. 2021.
\newblock On the sizes of openai api models.
\newblock \url{https://blog.eleuther.ai/gpt3-model-sizes/}.
\newblock Accessed: 2022-10-27.

\bibitem[{Gao et~al.(2020)Gao, Biderman, Black, Golding, Hoppe, Foster, Phang,
  He, Thite, Nabeshima et~al.}]{gao2020pile}
Leo Gao, Stella Biderman, Sid Black, Laurence Golding, Travis Hoppe, Charles
  Foster, Jason Phang, Horace He, Anish Thite, Noa Nabeshima, et~al. 2020.
\newblock The pile: An 800gb dataset of diverse text for language modeling.
\newblock \emph{arXiv preprint arXiv:2101.00027}.

\bibitem[{Gao et~al.(2021)Gao, Fisch, and Chen}]{gao2021making}
Tianyu Gao, Adam Fisch, and Danqi Chen. 2021.
\newblock Making pre-trained language models better few-shot learners.
\newblock In \emph{Proceedings of the 59th Annual Meeting of the Association
  for Computational Linguistics and the 11th International Joint Conference on
  Natural Language Processing (Volume 1: Long Papers)}, pages 3816--3830.

\bibitem[{Gulordava et~al.(2018)Gulordava, Bojanowski, Grave, Linzen, and
  Baroni}]{gulordava2018colorless}
Kristina Gulordava, Piotr Bojanowski, {\'E}douard Grave, Tal Linzen, and Marco
  Baroni. 2018.
\newblock Colorless green recurrent networks dream hierarchically.
\newblock In \emph{Proceedings of the 2018 Conference of the North American
  Chapter of the Association for Computational Linguistics: Human Language
  Technologies, Volume 1 (Long Papers)}, pages 1195--1205.

\bibitem[{Li et~al.(2022)Li, Cotterell, and Sachan}]{li2022probing}
Jiaoda Li, Ryan Cotterell, and Mrinmaya Sachan. 2022.
\newblock Probing via prompting.
\newblock In \emph{Proceedings of the 2022 Conference of the North American
  Chapter of the Association for Computational Linguistics: Human Language
  Technologies}, pages 1144--1157.

\bibitem[{Lin et~al.(2022)Lin, Mihaylov, Artetxe, Wang, Chen, Simig, Ott,
  Goyal, Bhosale, Du et~al.}]{lin2021few}
Xi~Victoria Lin, Todor Mihaylov, Mikel Artetxe, Tianlu Wang, Shuohui Chen,
  Daniel Simig, Myle Ott, Naman Goyal, Shruti Bhosale, Jingfei Du, et~al. 2022.
\newblock Few-shot learning with multilingual language models.
\newblock In \emph{Proceedings of the 2022 Conference on Empirical Methods in
  Natural Language Processing (EMNLP)}.

\bibitem[{Liu et~al.(2019)Liu, Gardner, Belinkov, Peters, and
  Smith}]{liu2019linguistic}
Nelson~F. Liu, Matt Gardner, Yonatan Belinkov, Matthew~E. Peters, and Noah~A.
  Smith. 2019.
\newblock Linguistic knowledge and transferability of contextual
  representations.
\newblock In \emph{Proceedings of the Conference of the North American Chapter
  of the Association for Computational Linguistics: Human Language
  Technologies}.

\bibitem[{Liu et~al.(2021{\natexlab{a}})Liu, Yuan, Fu, Jiang, Hayashi, and
  Neubig}]{liu2021pre}
Pengfei Liu, Weizhe Yuan, Jinlan Fu, Zhengbao Jiang, Hiroaki Hayashi, and
  Graham Neubig. 2021{\natexlab{a}}.
\newblock Pre-train, prompt, and predict: A systematic survey of prompting
  methods in natural language processing.
\newblock \emph{arXiv preprint arXiv:2107.13586}.

\bibitem[{Liu et~al.(2022{\natexlab{a}})Liu, Jiang, Monath, Cotterell, and
  Sachan}]{liu2022autoregressive}
Tianyu Liu, Yuchen Jiang, Nicholas Monath, Ryan Cotterell, and Mrinmaya Sachan.
  2022{\natexlab{a}}.
\newblock Autoregressive structured prediction with language models.
\newblock In \emph{Proceedings of the 2022 Conference on Empirical Methods in
  Natural Language Processing (EMNLP)}.

\bibitem[{Liu et~al.(2022{\natexlab{b}})Liu, Ji, Fu, Tam, Du, Yang, and
  Tang}]{liu2022p}
Xiao Liu, Kaixuan Ji, Yicheng Fu, Weng Tam, Zhengxiao Du, Zhilin Yang, and Jie
  Tang. 2022{\natexlab{b}}.
\newblock P-tuning: Prompt tuning can be comparable to fine-tuning across
  scales and tasks.
\newblock In \emph{Proceedings of the 60th Annual Meeting of the Association
  for Computational Linguistics (Volume 2: Short Papers)}, pages 61--68.

\bibitem[{Liu et~al.(2021{\natexlab{b}})Liu, Wang, Kasai, Hajishirzi, and
  Smith}]{liu2021probing}
Zeyu Liu, Yizhong Wang, Jungo Kasai, Hannaneh Hajishirzi, and Noah~A Smith.
  2021{\natexlab{b}}.
\newblock Probing across time: What does roberta know and when?
\newblock In \emph{Findings of the Association for Computational Linguistics:
  EMNLP 2021}, pages 820--842.

\bibitem[{Ma et~al.(2022)Ma, Zhou, Gui, Tan, Zhang, and Huang}]{ma2021template}
Ruotian Ma, Xin Zhou, Tao Gui, Yiding Tan, Qi~Zhang, and Xuanjing Huang. 2022.
\newblock Template-free prompt tuning for few-shot ner.
\newblock In \emph{Proceedings of the 2022 Conference of the North American
  Chapter of the Association for Computational Linguistics: Human Language
  Technologies}.

\bibitem[{Magar and Schwartz(2022)}]{magar2022data}
Inbal Magar and Roy Schwartz. 2022.
\newblock Data contamination: From memorization to exploitation.
\newblock In \emph{Proceedings of the 60th Annual Meeting of the Association
  for Computational Linguistics (Volume 2: Short Papers)}, pages 157--165.

\bibitem[{Min et~al.(2022)Min, Lyu, Holtzman, Artetxe, Lewis, Hajishirzi, and
  Zettlemoyer}]{min2022rethinking}
Sewon Min, Xinxi Lyu, Ari Holtzman, Mikel Artetxe, Mike Lewis, Hannaneh
  Hajishirzi, and Luke Zettlemoyer. 2022.
\newblock Rethinking the role of demonstrations: What makes in-context learning
  work?
\newblock In \emph{Proceedings of the 2022 Conference on Empirical Methods in
  Natural Language Processing (EMNLP)}.

\bibitem[{Nivre et~al.(2020)Nivre, de~Marneffe, Ginter, Hajic, Manning,
  Pyysalo, Schuster, Tyers, and Zeman}]{nivre2020universal}
Joakim Nivre, Marie-Catherine de~Marneffe, Filip Ginter, Jan Hajic,
  Christopher~D Manning, Sampo Pyysalo, Sebastian Schuster, Francis Tyers, and
  Daniel Zeman. 2020.
\newblock Universal dependencies v2: An evergrowing multilingual treebank
  collection.
\newblock In \emph{Proceedings of the 12th Language Resources and Evaluation
  Conference}, pages 4034--4043.

\bibitem[{Petroni et~al.(2019)Petroni, Rockt{\"a}schel, Riedel, Lewis, Bakhtin,
  Wu, and Miller}]{petroni2019language}
Fabio Petroni, Tim Rockt{\"a}schel, Sebastian Riedel, Patrick Lewis, Anton
  Bakhtin, Yuxiang Wu, and Alexander Miller. 2019.
\newblock Language models as knowledge bases?
\newblock In \emph{Proceedings of the 2019 Conference on Empirical Methods in
  Natural Language Processing and the 9th International Joint Conference on
  Natural Language Processing (EMNLP-IJCNLP)}, pages 2463--2473.

\bibitem[{Press et~al.(2022)Press, Zhang, Min, Schmidt, Smith, and
  Lewis}]{press2022measuring}
Ofir Press, Muru Zhang, Sewon Min, Ludwig Schmidt, Noah~A Smith, and Mike
  Lewis. 2022.
\newblock Measuring and narrowing the compositionality gap in language models.
\newblock \emph{arXiv preprint arXiv:2210.03350}.

\bibitem[{Raffel et~al.(2020)Raffel, Shazeer, Roberts, Lee, Narang, Matena,
  Zhou, Li, Liu et~al.}]{raffel2020exploring}
Colin Raffel, Noam Shazeer, Adam Roberts, Katherine Lee, Sharan Narang, Michael
  Matena, Yanqi Zhou, Wei Li, Peter~J Liu, et~al. 2020.
\newblock Exploring the limits of transfer learning with a unified text-to-text
  transformer.
\newblock \emph{J. Mach. Learn. Res.}, 21(140):1--67.

\bibitem[{Razeghi et~al.(2022)Razeghi, Logan~IV, Gardner, and
  Singh}]{razeghi2022impact}
Yasaman Razeghi, Robert~L Logan~IV, Matt Gardner, and Sameer Singh. 2022.
\newblock Impact of pretraining term frequencies on few-shot reasoning.
\newblock \emph{arXiv preprint arXiv:2202.07206}.

\bibitem[{Sang and Buchholz(2000)}]{sang2000introduction}
Erik Tjong~Kim Sang and Sabine Buchholz. 2000.
\newblock Introduction to the conll-2000 shared task chunking.
\newblock In \emph{Fourth Conference on Computational Natural Language Learning
  and the Second Learning Language in Logic Workshop}.

\bibitem[{Sang and De~Meulder(2003)}]{sang2003introduction}
Erik Tjong~Kim Sang and Fien De~Meulder. 2003.
\newblock Introduction to the conll-2003 shared task: Language-independent
  named entity recognition.
\newblock In \emph{Proceedings of the Seventh Conference on Natural Language
  Learning at HLT-NAACL 2003}, pages 142--147.

\bibitem[{Sanguinetti and Bosco(2014)}]{sanguinetti2014converting}
Manuela Sanguinetti and Cristina Bosco. 2014.
\newblock Converting the parallel treebank partut in universal stanford
  dependencies.
\newblock \emph{Converting the parallel treebank ParTUT in Universal Stanford
  Dependencies}, pages 316--321.

\bibitem[{Schick and Sch{\"u}tze(2021)}]{schick2021its}
Timo Schick and Hinrich Sch{\"u}tze. 2021.
\newblock It’s not just size that matters: Small language models are also
  few-shot learners.
\newblock In \emph{Proceedings of the 2021 Conference of the North American
  Chapter of the Association for Computational Linguistics: Human Language
  Technologies}, pages 2339--2352.

\bibitem[{Shi et~al.(2022)Shi, Suzgun, Freitag, Wang, Srivats, Vosoughi, Chung,
  Tay, Ruder, Zhou et~al.}]{shi2022language}
Freda Shi, Mirac Suzgun, Markus Freitag, Xuezhi Wang, Suraj Srivats, Soroush
  Vosoughi, Hyung~Won Chung, Yi~Tay, Sebastian Ruder, Denny Zhou, et~al. 2022.
\newblock Language models are multilingual chain-of-thought reasoners.
\newblock \emph{arXiv preprint arXiv:2210.03057}.

\bibitem[{Shliazhko et~al.(2022)Shliazhko, Fenogenova, Tikhonova, Mikhailov,
  Kozlova, and Shavrina}]{shliazhko2022mgpt}
Oleh Shliazhko, Alena Fenogenova, Maria Tikhonova, Vladislav Mikhailov,
  Anastasia Kozlova, and Tatiana Shavrina. 2022.
\newblock mgpt: Few-shot learners go multilingual.
\newblock \emph{arXiv preprint arXiv:2204.07580}.

\bibitem[{Silveira et~al.(2014)Silveira, Dozat, de~Marneffe, Bowman, Connor,
  Bauer, and Manning}]{silveira14gold}
Natalia Silveira, Timothy Dozat, Marie-Catherine de~Marneffe, Samuel Bowman,
  Miriam Connor, John Bauer, and Christopher~D. Manning. 2014.
\newblock A gold standard dependency corpus for {E}nglish.
\newblock In \emph{Proceedings of the Ninth International Conference on
  Language Resources and Evaluation (LREC-2014)}.

\bibitem[{Wang and Komatsuzaki(2021)}]{wang2021gpt-j}
Ben Wang and Aran Komatsuzaki. 2021.
\newblock {GPT-J-6B: A 6 Billion Parameter Autoregressive Language Model}.
\newblock \url{https://github.com/kingoflolz/mesh-transformer-jax}.

\bibitem[{Wolf et~al.(2019)Wolf, Debut, Sanh, Chaumond, Delangue, Moi, Cistac,
  Rault, Louf, Funtowicz et~al.}]{wolf2019huggingface}
Thomas Wolf, Lysandre Debut, Victor Sanh, Julien Chaumond, Clement Delangue,
  Anthony Moi, Pierric Cistac, Tim Rault, R{\'e}mi Louf, Morgan Funtowicz,
  et~al. 2019.
\newblock Huggingface's transformers: State-of-the-art natural language
  processing.
\newblock \emph{arXiv preprint arXiv:1910.03771}.

\bibitem[{Zeldes(2017)}]{zeldes2017gum}
Amir Zeldes. 2017.
\newblock \href {https://doi.org/http://dx.doi.org/10.1007/s10579-016-9343-x}
  {The {GUM} corpus: Creating multilayer resources in the classroom}.
\newblock \emph{Language Resources and Evaluation}, 51(3):581--612.

\bibitem[{Zeman et~al.(2017)Zeman, Popel, Straka, Hajic, Nivre, Ginter,
  Luotolahti, Pyysalo, Petrov, Potthast et~al.}]{zeman2017conll}
Daniel Zeman, Martin Popel, Milan Straka, Jan Hajic, Joakim Nivre, Filip
  Ginter, Juhani Luotolahti, Sampo Pyysalo, Slav Petrov, Martin Potthast,
  et~al. 2017.
\newblock Conll 2017 shared task: Multilingual parsing from raw text to
  universal dependencies.
\newblock In \emph{CoNLL 2017 Shared Task: Multilingual Parsing from Raw Text
  to Universal Dependencies}, pages 1--19. Association for Computational
  Linguistics.

\bibitem[{Zhang et~al.(2022)Zhang, Roller, Goyal, Artetxe, Chen, Chen, Dewan,
  Diab, Li, Lin et~al.}]{zhang2022opt}
Susan Zhang, Stephen Roller, Naman Goyal, Mikel Artetxe, Moya Chen, Shuohui
  Chen, Christopher Dewan, Mona Diab, Xian Li, Xi~Victoria Lin, et~al. 2022.
\newblock {OPT}: Open pre-trained transformer language models.
\newblock \emph{arXiv preprint arXiv:2205.01068}.

\bibitem[{Zhou et~al.(2022)Zhou, Sch{\"a}rli, Hou, Wei, Scales, Wang,
  Schuurmans, Bousquet, Le, and Chi}]{zhou2022least}
Denny Zhou, Nathanael Sch{\"a}rli, Le~Hou, Jason Wei, Nathan Scales, Xuezhi
  Wang, Dale Schuurmans, Olivier Bousquet, Quoc Le, and Ed~Chi. 2022.
\newblock Least-to-most prompting enables complex reasoning in large language
  models.
\newblock \emph{arXiv preprint arXiv:2205.10625}.

\end{thebibliography}
\bibliographystyle{acl_natbib}

\appendix

\section{Further Ablations and Analysis}
\label{app:analysis}
In this section, we test additional factors that may affect the performance of our proposed method. 

%Additional Error Analysis Results for Sec. 4.2
\begin{figure*}
    \begin{center}
        \subfloat[]{\includegraphics[height=23ex]{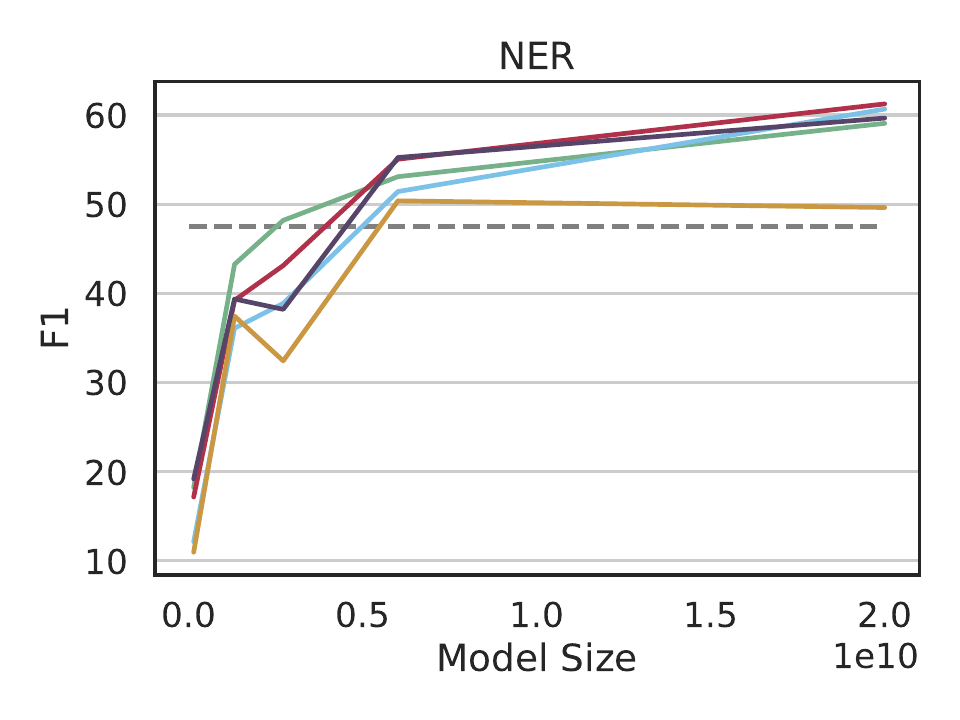}}
        \subfloat[]{\includegraphics[height=23ex]{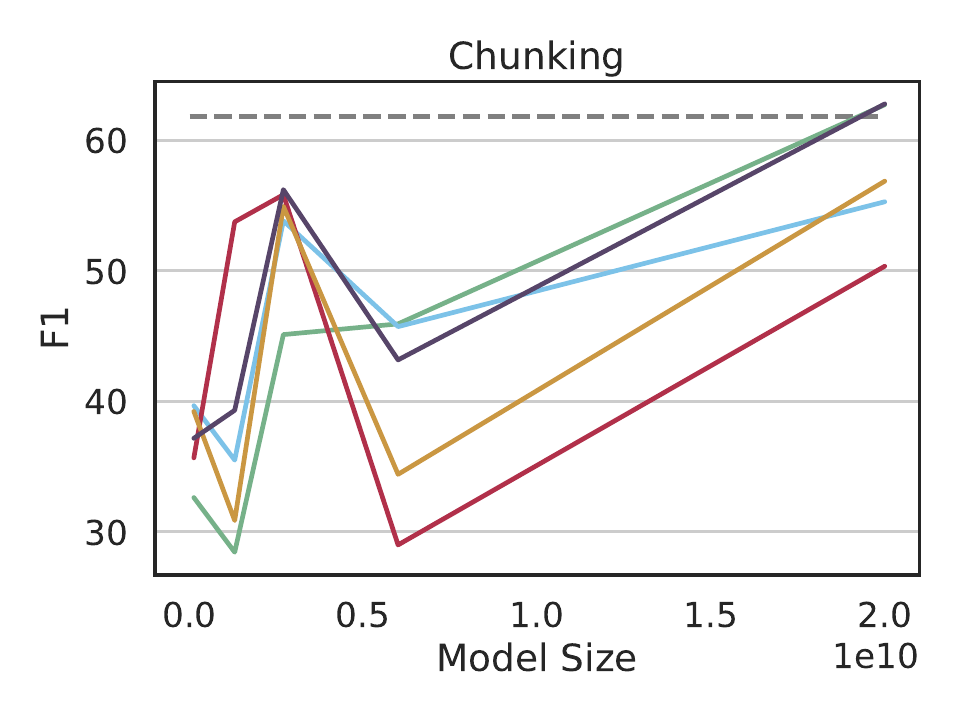}}
        \subfloat[]{\includegraphics[height=23ex]{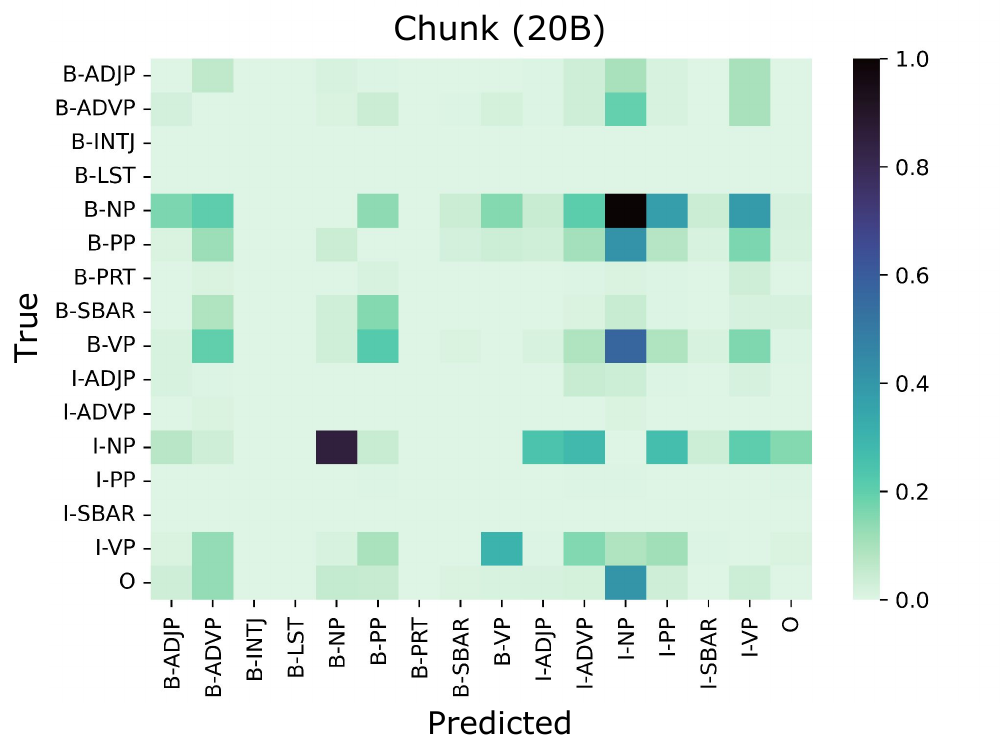}}
    \end{center}
    \caption{Additional error analysis results for Section \ref{sec:error-analysis}: performance across model sizes for different demonstrations sets on (a) NER and (b) chunking, and (c) confusion matrix for GPT-NeoX on chunking.}
    \label{fig:app-error-analysis}
\end{figure*}

\begin{table}[]
    \small
    \centering
    \begin{tabular}{c l @{\hspace{2.9\tabcolsep}} r r}
        \toprule
        \multirow{2}{*}{Task} & \multicolumn{1}{c}{\multirow{2}{*}{Model}} & \multicolumn{2}{c}{Eval Setting}\\
        & & \multicolumn{1}{c}{Fixed} &\multicolumn{1}{c}{Varied} \\
        \toprule
        \multirow{3}{*}{\makecell{POS\\(Acc.)}} & GPT-Neo-125M & 64.35 ± 1.6 & 64.38 ± 1.6 \\
        & GPT-Neo-2.7B & 70.36 ± 3.0 & 70.32 ± 3.0 \\
        & GPT-J-6B & 83.13 ± 1.1 & 83.10 ± 1.1 \\
        \hline 
        \multirow{3}{*}{\makecell{NER\\(F1)}} & GPT-Neo-125M & 16.03 ± 1.7 & 16.63 ± 2.1 \\
        & GPT-Neo-2.7M & 38.90 ± 2.7 & 38.72 ± 2.6 \\
        & GPT-J-6B & 51.43 ± 0.7 & 52.10 ± 0.9 \\
        \toprule
    \end{tabular}
    \caption{Results of ablating the choice of evaluation data for structured prompting on POS tagging and NER.}
    \label{tab:app_eval_ablation}
\end{table}

\subsection{Choice of evaluation set}
\label{app:analysis-eval}
For computational reasons, the models are evaluated on a fixed subset of 1000 randomly sampled test examples for each task. As using a smaller evaluation set can introduce noise into our performance estimates, we run a similar experiment on a number of the smaller models but resample the evaluation examples across five runs in addition to varying the demonstrations (Table \ref{tab:app_eval_ablation}). We find that varying the evaluation examples has a minimal effect on both the average performance and standard error on both POS tagging and NER.

\begin{table}[]
    \small
    \centering
    \begin{tabular}{c l @{\hspace{2.9\tabcolsep}} r r}
        \toprule
        \multirow{2}{*}{Task} & \multicolumn{1}{c}{\multirow{2}{*}{Model}} & \multicolumn{2}{c}{With BIO Constraints?}\\
        & & \multicolumn{1}{c}{Yes} &\multicolumn{1}{c}{No} \\
        \toprule
        \multirow{3}{*}{\makecell{NER\\(F1)}} & GPT-Neo-125M & 15.52 ± 1.7 & 16.03 ± 1.8 \\
        & GPT-J-6B & 53.03 ± 1.0 & 51.43 ± 0.7 \\
        & GPT-NeoX-20B & 58.05 ± 2.1 & 57.00 ± 1.9 \\
        \hline 
        \multirow{3}{*}{\makecell{Chunk\\(F1)}} & GPT-Neo-125M & 36.85 ± 1.3 & 38.32 ± 1.5 \\
        & GPT-J-6B & 39.63 ± 3.4 & 40.12 ± 3.5 \\
        & GPT-NeoX-20B & 57.60 ± 2.4 & 59.25 ± 2.7 \\
        \toprule
    \end{tabular}
    \caption{Results of ablating the BIO constraints for structured prompting on NER and chunking.}
    \label{tab:app_bio_ablation}
\end{table}

\subsection{Ablating BIO Constraints}
\label{app:analysis-bio}

During this work, we found that limiting the potential output tag space from the model with global BIO constraints made little difference in model performance for both NER and chunking (Table \ref{tab:app_bio_ablation}). Specifically, in every case, the difference between the two settings was within the standard error of the means across runs, with NER performing slightly better with the constraints and chunking performing slightly worse.

\begin{table*}[]
    \centering
    \begin{tabular}{c r |r r r}
        \toprule
        \multirow{2}{*}{\textbf{Model Size}} & \multirow{2}{*}{\textbf{k =}} & \multicolumn{3}{c}{\textbf{Task}} \\
         & & \textbf{POS} (Acc.) &\textbf{NER} (F1) & \textbf{Chunk} (F1) \\
        \hline
        125M & \multirow{5}{*}{10} & 64.35 ± 1.6 & 15.52 ± 1.7 & 36.85 ± 1.3 \\
        1.3B & & 68.45 ± 1.7 & 39.07 ± 1.2 & 37.56 ± 4.5 \\
        2.7B & & 70.36 ± 3.0 & 40.16 ± 2.6 & 53.18 ± 2.1 \\
        6B & & 83.13 ± 1.1 & 53.03 ± 1.0 & 39.63 ± 3.4 \\
        20B & & 83.56 ± 0.8 & 58.05 ± 2.1 & 57.60 ± 3.4 \\
        \hline 
        \multirow{5}{*}{20B} & 0 & 17.20 & 3.79 & 1.08 \\
        & 1 & 70.84 ± 1.9 & 10.26 ± 1.1 & 32.02 ± 3.9 \\
        & 3 & 79.08 ± 1.1 & 33.63 ± 2.8 & 48.33 ± 3.6 \\
        & 5 & 81.72 ± 1.2 & 40.60 ± 1.6 & 50.98 ± 3.0 \\
        & 7 & 82.67 ± 0.8 & 52.12 ± 3.7 & 54.00 ± 2.7 \\
        & 9 & 83.56 ± 0.8 & 58.08 ± 1.8 & 54.84 ± 2.9 \\
        \hline
        \hline 
        \multicolumn{5}{l}{\textbf{Baselines}} \\
        \hline
        \multicolumn{2}{c|}{Majority Label} & 17.75 & 0.00 & 0.00 \\
        \multicolumn{2}{c|}{Per-Span Majority} & 80.76 & 47.52 & 61.84 \\
        \toprule
    \end{tabular}
    \caption{Full Results of GPT-Neo series experiments from Section \ref{sec:overall-results}.}
    \label{tab:app_full_results}
\end{table*}

\subsection{Full Results of Error Analysis}
\label{app:error-analysis}
We provide additional error analysis results from Section \ref{sec:error-analysis}
in Figure \ref{fig:app-error-analysis}.

\subsection{Full Results of Pretraining Data Analysis}
\label{app:data-analysis}
The complete data analysis for labels not shown in Section \ref{sec:data-analysis} is detailed in Table \ref{tab:app_data_analysis}.

\section{Complete Results of Structured Prompting Experiments}
\label{app:all-results}
We provide the full numerical results for the experiments in Section \ref{sec:overall-results} in Table \ref{tab:app_full_results}. 

\section{Responsible NLP Miscellanea}
This section details information from the Responsible NLP Checklist not covered elsewhere in the paper. 

\paragraph{Compute Costs}
The computational cost of each prompting experiment on the GPT-Neo series of models varies depending on the task and size of the underlying PLM: run times for a single experiment range from around 43 minutes for POS tagging on the 125M parameter model to approximately 50 hours for chunking with GPT-NeoX (20B parameters). The smaller GPT-neo models (fewer than 6B parameters) are run on a single Nvidia RTX-6000, and larger models are run on one or more Nvidia A40 GPUs. 

For the GPT-3 POS tagging experiments, we run the models through the OpenAI API. When performing constrained decoding through the API, each example requires multiple calls per word in the sentence to decode the label forms, since model state caching for custom decoding is not available. For GPT-Curie (k=5), with constrained decoding, on average 230M tokens are submitted to the API per run; with Davinci (k=10, where we only performed unconstrained decoding), an average of 1.2M tokens are submitted per run. 

%Data Analysis Results Table (Appendix
\begin{table}[]
    \centering
    \small
    \begin{tabular}{l|r r}
        \toprule
        \textbf{Label} & \textbf{Freq.} & \textbf{Task Stats} \\
        \hline
        \hline
        \multicolumn{2}{l}{\textbf{POS Tagging}} & \textbf{UD Format} \\
        \hline
        ADJ & 449,789 & 2.49\% \\
        ADP & 1,847,009 & 0.80\% \\
        ADV & 2,315,004 & 0.42\% \\
        AUX & 572,373 & 1.71\% \\
        CCONJ & 22,050 & 23.48\% \\
        DET & 1,528,722 & 0.72\% \\
        INTJ & 28,882 & 2.11\% \\
        NOUN & 360,034 & 9.29\% \\
        NUM & 3,642,199 & 0.10\% \\
        PART & 4,573,194 & 0.09\% \\
        PRON & 130,754 & 11.00\% \\
        PROPN & 50,247 & 18.81\% \\
        PUNCT & 131,344 & 18.27\% \\
        SCONJ & 18,307 & 17.68\% \\
        SYM & 1,189,552 & 0.08\% \\
        VERB & 451,447 & 4.66\% \\
        X & -- & -- \\
        \hline
        \hline
        \multicolumn{3}{l}{\textbf{NER}}
        \\
        \hline
        B-PER & 5,655 & -- \\
        I-PER & 4,678 & -- \\
        B-ORG & 2,603 & -- \\
        I-ORG & 3,793 & -- \\
        B-LOC & 4,467 & -- \\
        I-LOC & 2,197 & -- \\
        B-MISC & 1,133 & -- \\
        I-MISC & 907 & -- \\
        O & -- & -- \\
        \toprule
    \end{tabular}
    \caption{Automatic analysis of the Pile for labels from UD POS tagset and CONLL03 NER tagset. \textit{Task Stats} document the percentage of occurrences that are in the UD format for POS tagging. We do not search labels that are individual characters due to how frequently they appear in the corpus.}
    \label{tab:app_data_analysis}
\end{table}

\paragraph{Intended Usage of Artifacts} 
To the best of our knowledge, our experiments all fall within the intended use cases of the GPT-Neo models and the Pile dataset, as well as the usage policy of the OpenAI API.

\end{document}